\documentclass[sigconf]{aamas} 

\usepackage{preamble}

\newif\ifarxiv
\arxivtrue

\acmSubmissionID{571}

\title[\coach: Enhancing Human Teamwork through AI Coaching]{\coach: Enhancing Human Teamwork via AI-enabled Coaching}

\author{Sangwon Seo}
\affiliation{
  \institution{Rice University}
  \city{Houston, TX}
  \country{USA}}
\email{sangwon.seo@rice.edu}

\author{Bing Han}
\affiliation{
  \institution{Rice University}
  \city{Houston, TX}
  \country{USA}}
\email{bing.han@rice.edu}

\author{Rayan E. Harari}
\affiliation{
  \institution{Harvard Medical School}
  \city{Boston, MA}
  \country{USA}}
\email{rharari@bwh.harvard.edu}

\author{Roger D. Dias}
\affiliation{
  \institution{Harvard Medical School}
  \city{Boston, MA}
  \country{USA}}
\email{rdias@bwh.harvard.edu}

\author{Marco A. Zenati}
\affiliation{
  \institution{Harvard Medical School}
  \city{Boston, MA}
  \country{USA}}
\email{marco_zenati@hms.harvard.edu}

\author{Eduardo Salas}
\affiliation{
  \institution{Rice University}
  \city{Houston, TX}
  \country{USA}}
\email{eduardo.salas@rice.edu}

\author{Vaibhav Unhelkar}
\affiliation{
  \institution{Rice University}
  \city{Houston, TX}
  \country{USA}}
\email{vaibhav.unhelkar@rice.edu}

\begin{abstract}
Coaches are vital for effective collaboration, but cost and resource constraints often limit their availability during real-world tasks. This limitation poses serious challenges in life-critical domains that rely on effective teamwork, such as healthcare and disaster response. To address this gap, we propose and realize an innovative application of AI: task-time team coaching. Specifically, we introduce \coach, a novel AI system that complements human coaches by providing real-time guidance during task execution. \coach monitors team behavior, detects misalignments in team members' shared understanding, and delivers automated interventions to improve team performance. We validated \coach through two human subject experiments involving dyadic collaboration. The results demonstrate that the system significantly enhances team performance with minimal interventions. Participants also perceived \coach as helpful and trustworthy, supporting its potential for adoption. Our findings also suggest promising directions both for AI research and its practical applications to enhance human teamwork.
\end{abstract}



\begin{document}

%%% The following commands remove the headers in your paper. For final 
%%% papers, these will be inserted during the pagination process.

\pagestyle{fancy}
\fancyhead{}

%%% The next command prints the information defined in the preamble.

\maketitle 

%%%%%%%%%%%%%%%%%%%%%%%%%%%%%%%%%%%%%%%%%%%%%%%%%%%%%%%%%%%%%%%%%%%%%%%%

\section{Introduction}
\label{sec. intro}
Consider your favorite sports team --- whether it is soccer, cricket, basketball, or another team sport --- working together to achieve a common goal. Even though all the team members are trained professionals, some teams consistently outperform others. Indeed, \textit{a team of individual experts does not necessarily make for an expert team}~\cite{bisbey2021transforming}; building a successful team requires the confluence of multiple factors~\cite{mathieu2000influence}. Human factors research has identified key drivers of team effectiveness, including capability, coordination, communication, and coaching~\cite{tannenbaum2020teams}. Through targeted training and interventions, human teams can significantly improve coordination and enhance their performance in collaborative tasks.
\ifarxiv
\blfootnote{This article is an extended version of an identically-titled paper accepted at the \textit{International Conference on Autonomous Agents and Multiagent Systems (AAMAS 2025)}.}
\else
\blfootnote{An extended version of this paper, which includes supplementary material mentioned in the text, is available at \url{http://tiny.cc/socratic-appendix}}
\fi

Coaches play a crucial role in both team training and interventions. Rather than performing tasks themselves, they enhance collaboration by offering expert insights. These insights are provided both during task execution, such as in games, and during training sessions, such as in practice. While coaches are common in professional sports, integrating them into life-critical fields presents significant challenges~\cite{seo2021towards, orlov2024rw4t}. Resource constraints and a shortage of experts make it difficult to employ coaches during task execution.
For example, in surgical teamwork, a coach could be invaluable in reducing preventable medical errors~\cite{wahr2013patient, makary2016medical, seo2021towards}. Reducing these errors would significantly improve patient health outcomes. However, due to the shortage of medical professionals, it is not feasible for a specialist to continuously serve in this coaching role. Similarly, in aviation, coaches assist with simulation-based training, but they cannot accompany a flight crew on every flight~\cite{kolander2019flight}. 

\begin{figure*}[t]
  \centering
  \includegraphics[width=0.9\linewidth]{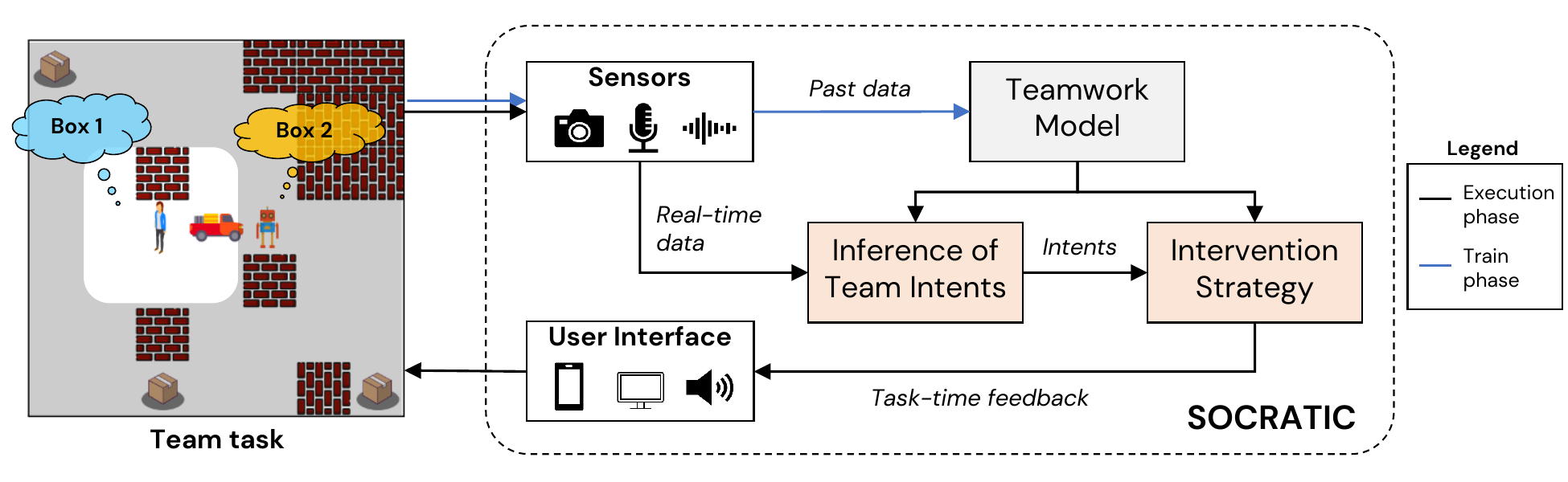}
  \caption{Schematic of \coach: an AI coach for enhancing teamwork during task execution. Blue arrows represent the workflow during the training phase, whereas black arrows indicate the workflow during the execution phase.} 
  \label{fig. schematic} 
  \Description{A schematic diagram that provides an overview of \coach.}
\end{figure*}

Recognizing the need for coaching assistance in life- and safety-critical applications, we propose an innovative use of artificial intelligence (AI): task-time team coaching. Specifically, we envision an AI agent that complements a human coach by monitoring a team during task execution and providing real-time guidance to improve teamwork, particularly in situations where the human coach may be busy or unavailable. While coaches offer a variety of feedback before, during, and after tasks, in this work, we limit our scope to delivering task-time feedback in time-critical tasks. For this setting, we present a novel proof-of-concept AI agent, \coach, designed to complement a human coach and thereby enhance teamwork.

Illustrated in \cref{fig. schematic}, the overall design of \coach is grounded in the extensive literature on human team training. Specifically, \coach operates by observing task execution and identifying points where the team's mental models regarding shared plans may become misaligned. When such a misalignment is detected, \coach prompts the team to pause, reflect on their plans, and offers suggestions for improvement. By encouraging the team to reconsider future actions that could lead to inefficiencies or errors, \coach aims to enhance collaborative decision-making. 

From an AI perspective, \coach leverages recent advances in imitation learning and multi-agent systems. First, it employs multi-agent imitation learning to model team behavior based on demonstrations from previously executed tasks. Using this model and data from an ongoing task, it builds on \tic -- a recent algorithm for agent-based teamwork -- to algorithmically detect points of misalignment and generate recommendations. Finally, through an interactive user interface, \coach delivers the automatically generated interventions aimed at aligning the team's understanding and improving overall performance.

To evaluate \coach, we conducted two human subject experiments: one focused on training and the other on validation. Both experiments involved two collaborative tasks and dyadic teams. In the training experiment, we curated a novel dataset of human demonstrations annotated with intents and used it to train \coach. In the validation experiment, we conducted a randomized controlled trial to evaluate both \coach's objective performance and the users' subjective perceptions of the system. The experimental results show that Socratic significantly improves team performance with minimal interventions. Equally important for its adoption, participants perceive \coach as helpful to improving teamwork. The evaluations also suggest promising directions for both AI research and the proposed applications, highlighting the potential of AI agents to support human teamwork.

\section{Background}
\label{sec: related-work}
Before describing \coach, we present the concepts and related research that inform our approach. 

\begin{figure*}[h]
  \centering
  \begin{subfigure}[b]{0.245\linewidth}
      \centering
      \includegraphics[width=0.95\textwidth, frame]{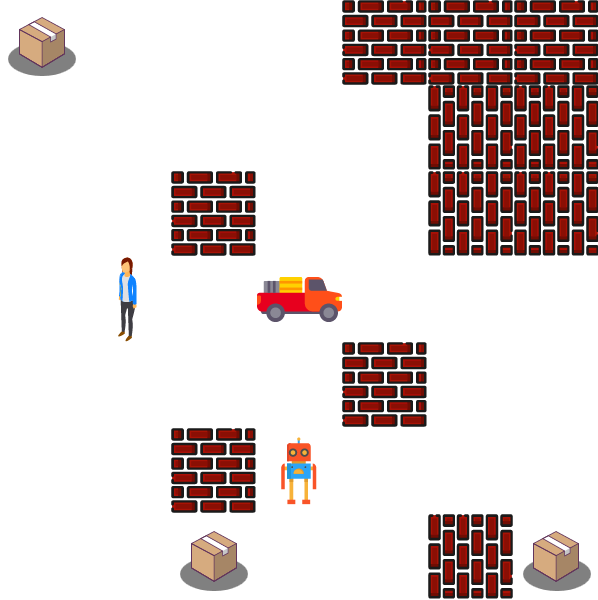}
      \caption{Map of \movers}
      \label{fig: movers}
  \end{subfigure}
  \hfill  
  \begin{subfigure}[b]{0.245\linewidth}
      \centering
      \includegraphics[width=0.95\textwidth, frame]{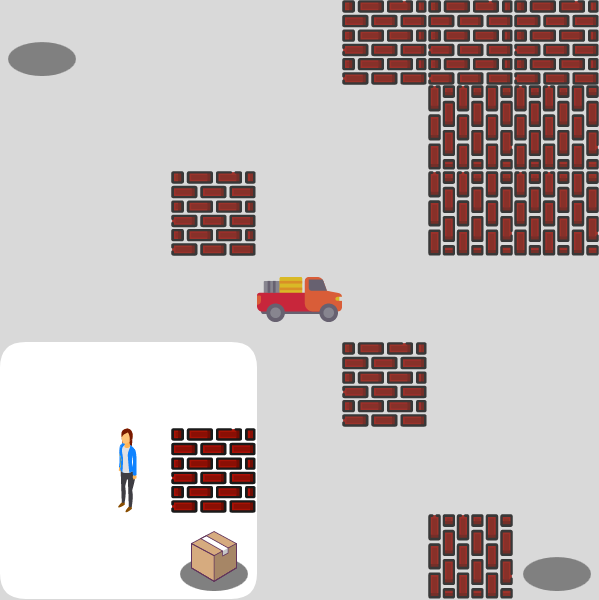}
      \caption{\movers: Alice's perspective}
      \label{fig: movers human sight}
  \end{subfigure}
  \hfill  
  \begin{subfigure}[b]{0.245\linewidth}
      \centering
      \includegraphics[width=0.95\textwidth, frame]{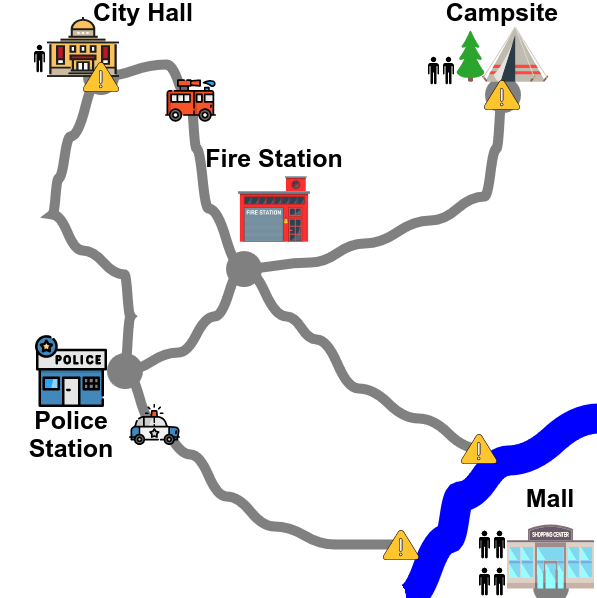}
      \caption{Map of \rescue}
      \label{fig: rescue}
  \end{subfigure}
  \hfill  
  \begin{subfigure}[b]{0.245\linewidth}
      \centering
      \includegraphics[width=0.95\textwidth, frame]{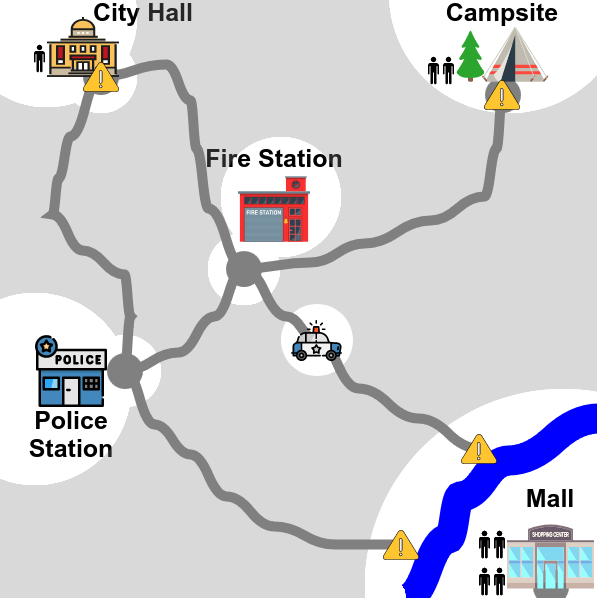}
      \caption{\rescue: Police's perspective}
      \label{fig: rescue human sight}
  \end{subfigure}

  \captionsetup{subrefformat=parens}
  \caption{\movers and \rescue domains, detailed in \cref{sec. domains}. Team members can observe only the unshaded region of the environment.}
  \label{fig: domains}
\end{figure*}
\subsection{Collaborative Tasks}
\label{sec. task model}
Teamwork is fundamental to many human endeavors, spanning scenarios such as sports, healthcare, aviation, and more. Our focus is on \textit{time-critical scenarios}, such as healthcare and disaster response, where effective teamwork is crucial for mission success. Teamwork occurs at various levels, ranging from large organizations to small ad-hoc teams. We focus on \textit{mission-oriented, sequential tasks}, where an established team works toward a clearly defined mission (e.g., \cref{fig: domains}). Although the mission is well-defined, there are often multiple ways to achieve the task. Real-world challenges, such as uncertainty, information asymmetry, and partial observability, can create barriers to efficient teamwork and task completion. Finally, we consider teams composed of human members, either in human-only teams or hybrid human-AI teams.

To develop an AI agent capable of supporting such teamwork, the first step is to mathematically model the task and team dynamics. Fortunately, research in multi-agent systems offers several established formalisms for modeling collaborative tasks, including belief-desire-intention frameworks, Markov models, and game theory~\cite{tambe2005conflicts, bernstein2002complexity, nair2003taming, nair2005hybrid, chao2016timed, stone2010ad, grosz1999planning, grosz1996collaborative, modi2005adopt, harbers2012measuring, semsar2009multi, fridovich2020efficient, tian2022safety}. In our work, we leverage decentralized multi-agent partially observable Markov decision processes (Dec-POMDPs)~\cite{oliehoek2016concise}. This choice is motivated by their ability to model tasks with well-defined missions, structured teams, time constraints, action uncertainties, and partial observability, as well as their prior use in modeling time-critical collaboration scenarios like disaster response~\cite{chen2015decentralized, unhelkar2016contact, liu2017learning, lee2021multi,  dong2023optimizing}. 

We define a task as the tuple $\mathcal{M} = (n, S, A, \Omega, T, O, R, \gamma, h)$, where $n$ is the number of agents, $S$, $A\doteq\times_i A_i$ and $\Omega\doteq\times_i\Omega_i$ denote a state space, an action space and an observation space, respectively, $T(s'|s, a)$ denotes a probability of a state $s$ transitioning to another state $s'$ given a joint action $a\doteq(a_1, \cdots, a_n)$, $O(o|s', a)$ is a probability of a joint observation $o\doteq(o_1, \cdots, o_n)$ given a state $s'$ and a joint action $a$, $R$ is the task reward (objective), $\gamma$ is the discount factor, and $h$ is the task horizon. In theory, a team could act optimally by computing a decentralized policy using Dec-POMDP solvers based on the task model. However, it is unrealistic to expect human team members to compute and execute such a policy flawlessly and without errors. Therefore, we draw on human factors research to model team behaviors and identify strategies for improving teamwork.

\subsection{The Science of Human Teamwork}
\label{sec: team training}

% \subsubsection{Drivers of Effective Teamwork}
The science of human teamwork focuses on the question: \textit{What makes teams work?}~\cite{salas2018science}.
Over the past four decades, psychologists and human factors researchers have systematically identified the factors that make teamwork challenging and developed methods to improve it~\cite{salas2008teams,cooke2013interactive,cooke2021effective,salas2024science}.
We briefly review key insights that inform our work, while directing readers to recent survey by ~\citet{tannenbaum2020teams} for more details.
There is broad consensus that teamwork is especially challenging in time-critical scenarios, where success depends on the convergence of multiple factors.
A major challenge is that humans often make suboptimal decisions due to bounded rationality~\cite{simon1997models, kahneman2003maps, kahneman2013prospect} and limited situational awareness~\cite{endsley2000situation, parush2011communication, stanton2017state, endsley2021situation}, especially under time constraints.
Hence, \coach does not assume perfect rationality or situational awareness from team members.
Even teams composed of experts may not function optimally due to a lack of shared mental models, leading to poor coordination and even fatal errors~\cite{cannon1993shared, mathieu2000influence, jonker2010shared, van2011team, mccomb2014concept, harari2024misalignment}.
Thus, \coach explicitly considers team members' intent and allows for potential misalignment, which can lead to suboptimal teamwork.

% \subsubsection{Team Training}
To enhance teamwork, the science of teamwork recommends several methods and best practices, including effective communication, simulation-based training, and coaching -- the latter being the focus of this paper. Coaches play a crucial role by assessing teamwork and providing feedback to improve it. While human coaches rely on their expertise and experience for these activities, the science of teamwork has developed principled methods and formalized best practices for coaching. Researchers have established robust methods for assessing  teams~\cite{salas2018science,granaasen2019towards,costar2020improving,grimm2023dynamical,kennedy2024novel} and generating targeted insights to enhance teamwork~\cite{hackman2005theory, salas2008does, peters2013team, weaver2014team, britton2015expanding}. However, these assessments are typically post-hoc, lack automation, and are limited to contexts where a human coach is available. Thus, we explore the design of an AI coach capable of operationalizing these insights, detecting misalignments in team members' shared intents, and providing real-time feedback during task execution. 

\subsection{AI-Assisted Teamwork}
AI-assisted human teamwork is an emerging area of research with applications being explored across various domains~\cite{kamar2009incorporating, rajvsp2020systematic, seo2021towards, pynadath2023effectiveness, pynadath2023improving, van2023markov, seo2024ai, harari2024deep}. For instance, DeepMind and Liverpool FC are investigating data-driven approaches to analyze and enhance team strategies in football~\cite{tuyls2021game}. For applications in healthcare and disaster response, researchers have applied AI to analyze team conversations and improve extended-duration teamwork~\cite{kim2016improving, amir2016mutual}. Closer to our focus on time-critical scenarios, domain-specific methods for automated teamwork assessment have been developed~\cite{granaasen2019towards, kotlyar2023assessing}. However, these methods, to our knowledge, provide only post-hoc support, and AI has not yet been used for task-time coaching.

Approaches for assessing and improving teamwork in human-robot or robot-only teams are also relevant to our work~\cite{riley2004advice, riley2002empirical, wu2022evaluating, thomaz2016computational, unhelkar2016contact}. Research in human-robot collaboration introduces metrics for evaluating teamwork~\cite{hoffman2019evaluating, ma2022metrics, norton2022metrics} and algorithms for improving it~\cite{buisan2021human, semeraro2023human, mukherjee2022survey, nikolaidis2017human, unhelkar2020decision}. However, these methods focus on training robots to work with humans. In contrast, our work centers on an AI agent that provides coaching and decision support, without directly performing the task.
Closest to our work are the recent frameworks TIC~\cite{seo2023automated} and TARS~\cite{zhang2024risk}, which generate task-time interventions to enhance multi-agent teamwork. TARS uses Dynamic Epistemic Logic-POMDP to generate interventions through planning algorithms~\cite{zhang2024risk}. TIC employs Dec-POMDPs and multi-agent imitation learning to generate interventions through a learned model~\cite{seo2023automated,seo2022semi}. However, these methods have not been applied or evaluated in settings with human team members.

Our work builds on these methods but differs in key ways. First, we adopt a systems perspective to develop \coach that includes both an intervention algorithm and a user interface, enabling interaction with and coaching for human users. Second, our methodology incorporates mechanisms to collect training data on human teamwork, including their cognitive states. Finally, we validate the effectiveness of the solution through human subject experiments.

\section{\coach}
\label{sec. coach}
\begin{figure}[h]
  \centering
  \newcommand\gap{0.495}
  \begin{subfigure}[t]{\gap\linewidth}
      \centering
      \includegraphics[width=\textwidth]{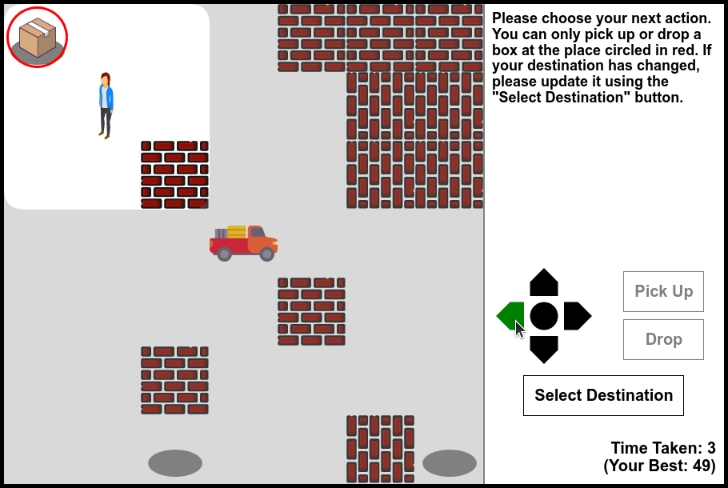}
      \caption{Task scene and UI for Alice.}
      \label{fig. ui for task}
  \end{subfigure}
  \hfill
    \begin{subfigure}[t]{\gap\linewidth}
      \centering
      \includegraphics[width=\textwidth]{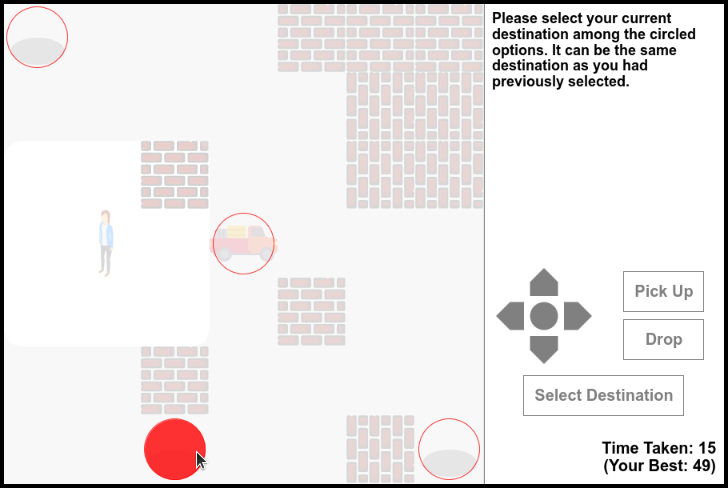}
      \caption{UI for intent annotation.}
      \label{fig. ui for intent selection}
  \end{subfigure}
  \hfill
  \begin{subfigure}[t]{\gap\linewidth}
      \centering
      \includegraphics[width=0.95\textwidth, frame]{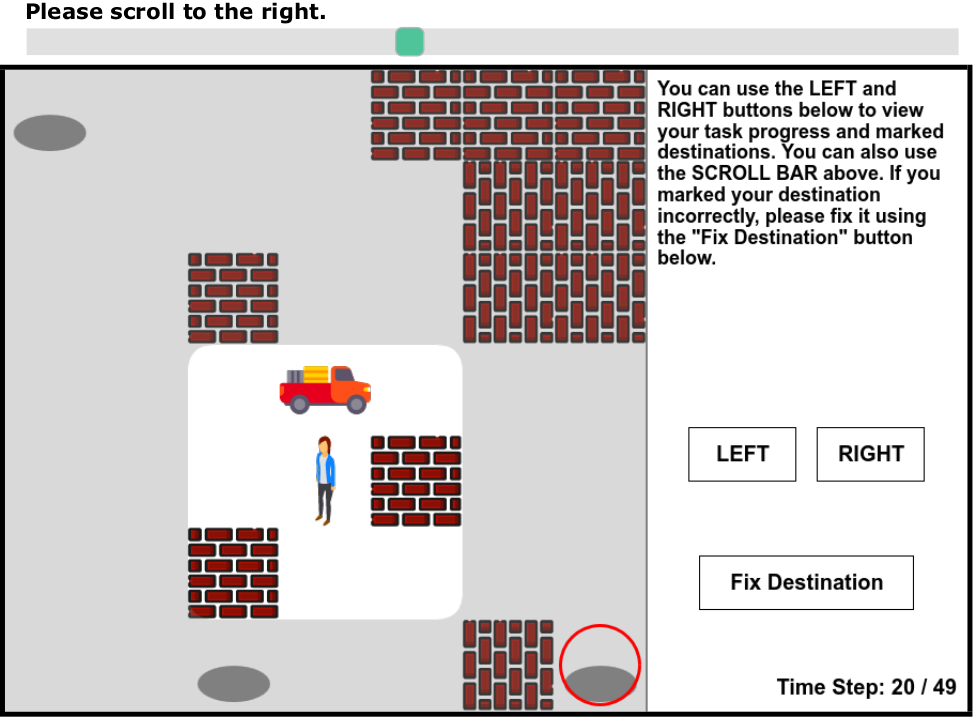}
      \caption{UI for after-action review.}
      \label{fig. ui for post-session review}
  \end{subfigure}
  \captionsetup{subrefformat=parens}
  \caption{Snapshots of the \movers task from the first study (larger images are available in the appendix).}
  \label{fig. ui}
\end{figure}
We now describe \coach: the \textit{System for Objective Coaching through Automated Task-time Interventions for Collaboration}.
Drawing on multiple disciplines (\cref{sec: related-work}), we begin by outlining the system's design requirements and architecture.
We then detail its key components: a module for monitoring team performance, an algorithm for learning team behavior models, another for generating task-time interventions, and a user interface to communicate these interventions to the team.
We illustrate \coach using two human-AI collaboration tasks, detailed in \cref{sec. domains} and inspired by real-world scenarios, implemented on a web-based simulation platform.

\subsection{System Overview}

\subsubsection{Scope}
We limit our scope to collaborative tasks modeled as Dec-POMDPs (\cref{sec. task model}) and teams that include at least one human member.
Importantly, we do not make assumptions about team members' rationality or expertise levels.
As reviewed in \cref{sec: team training}, the science of teamwork identifies several key drivers of effective teamwork.
In this proof-of-concept work, we focus on team alignment\footnote{Investigating other drivers of effective teamwork, intervention mechanisms, and teamwork settings is an important avenue for future research.} 
 --- ensuring that the team is ``on the same page.''
Misalignment is particularly common in time-critical scenarios, where teams may lack sufficient time to communicate and coordinate shared plans.
Additionally, real-world factors such as partial observability, fatigue, and uncertainty can further degrade team member's understanding of each other's beliefs, desires, and intentions.

\subsubsection{Design Requirements}
\label{sec: problem statement}
\label{sec. requirements}
With this scope defined, we design \coach: an AI-enabled coaching agent to improve teamwork during task execution. The design process began by identifying system requirements through brainstorming sessions with an interdisciplinary team of researchers in human factors, team training, AI, and usability. We determined that an AI agent capable of \textit{detecting misalignments in team members' intents} and \textit{alerting the team to pause, reflect, and adjust their plans} is both feasible to develop and can significantly enhance collaboration.
For the successful realization and adoption of such an agent, we distilled key requirements (\textbf{Rx}); namely, \coach must be:
\begin{itemize}
    \item[\textbf{R1.}] able to sense and monitor teamwork;
    \item[\textbf{R2.}] able to accurately infer intents of the team members;
    \item[\textbf{R3.}] able to accurately anticipate future actions of the team;
    \item[\textbf{R4.}] able to generate effective task-time interventions;
    \item[\textbf{R5.}] able to effectively deliver the interventions; and
    \item[\textbf{R6.}] perceived as useful by the team members.
\end{itemize}

\subsubsection{System Architecture}
To meet the design requirements, \coach leverages recent advancements in imitation learning and multi-agent systems, incorporating an interactive user interface to monitor the team and deliver interventions.
For \textbf{R1} (sensing and monitoring teamwork), we assume \coach is equipped with sensors to observe both the team and task environment. Similar to sport scenarios, where team members may have partial observability, the coach has full visibility of the environment.
To meet \textbf{R2} (inferring intents) and \textbf{R3} (anticipating future actions), \coach employs a recent multi-agent imitation learning algorithm \btil that explicitly models team members' intents and learns a generative model of team behavior~\cite{seo2022semi}.
Building on this model, \coach utilizes a specialized instance of the \tic framework to generate task-time interventions to meet \textbf{R4}~\cite{seo2023automated}.
Lastly, \coach includes a user interface to deliver these interventions to the team, addressing \textbf{R5}. 

\subsubsection{System Operation}
\coach operates in two phases: training and execution. During the training phase, \coach observes the team performing tasks in practice sessions, collecting teamwork data and learning generative models of team behavior. During task execution, \coach uses the learnt generative model to infer team intents, detect misalignments, and compute and deliver effective interventions. We now detail each system component.

\subsection{Training Phase}
\label{sec. tic training}

\subsubsection{Team Model}
\label{sec. teamwork model}
To effectively monitor the team, \coach builds upon a mathematical model of the task and team behavior. Having described the task model in \cref{sec. task model}, we now formalize the model of team behavior. Human decision-making often depends on factors beyond the task state, such as cognitive states corresponding to beliefs and intents~\cite{hiatt2017human, neubauer2020multimodal}. Hence, \coach explicitly models the influence of team members' intent -- a latent variable -- on their behavior. More specifically, following the Agent Markov Model (AMM)~\cite{unhelkar2019learning}, $j$-th team member's behavior is defined by the tuple $\mathcal{H}_j = (X, \pi_j, \zeta_j; \mathcal{M})$, where $X$ represents the set of possible task-specific intents, $\pi_j(a|x, s)$ denotes the team member's policy, and $\zeta_j(x'|s', a, x)$ represents the intent transition model.\footnote{%
Although team members' behavior may also depend on other latent factors, such as cognitive states and beliefs about unobserved parts of the environment, the decision to model only intent simplifies the system design. Our experiments confirm that this modeling choice is valid for the domains considered. However, we believe that performance of future AI-enabled coaching systems could be further enhanced by incorporating additional decision factors and more sophisticated behavioral models.}
While this model is well-defined, it is not trivial for domain experts to specify. Therefore, \coach leverages imitation learning to learn the model parameters from demonstrations collected during training sessions.

\subsubsection{Model Learning}
In particular, \coach uses \btil to learn the unknown parameters of the team behavioral model: $\pi(a|s, x)$ and $\zeta(x'|s', x, a)$.
\btil is a multi-agent imitation learning algorithm that explicitly models latent decision factors, such as intents~\cite{seo2022semi}.
By leveraging a Bayesian approach, \btil has been shown to attain sample- and label- efficient model learning from team demonstrations.
Additionally, \btil can learn from both optimal and sub-optimal demonstrations.
This is especially important for \coach, as it learns the team model from demonstrations collected during practice sessions, where team behavior may not always be optimal.
In practice, \coach's model learning begins with the collection of data on observable features of team demonstration, specifically $(s,a)$-trajectories.
With the assistance of a human annotator, a subset of these trajectories is annotated with the values of team intent $(x)$.
Using this combination of trajectory data and intent annotations, \coach utilizes the semi-supervised variant of \btil to learn the team behavioral model $\mathcal{H}_j \forall j = 1:n$.

\subsubsection{Team Monitoring}
Equally critical to team modeling are the mechanisms for monitoring the team and collecting teamwork data: specifically, $(s,a)$-trajectories and annotations of $(x)$ for a subset of the training data. In this proof-of-concept, we focus on collaborative tasks conducted through a web-based interface and develop methods for data collection and annotation specific to this setting, illustrated in \cref{fig. ui} and detailed in \cref{sec: data collection}. For real-world applications, we recommend using multimodal sensors to monitor and gather teamwork data. We leave the exploration of related perception challenges for future work, with relevant research directions discussed in \cref{sec: conclusion}. \coach uses the same monitoring mechanisms during the task execution phase, which we describe next.

\begin{figure}[t]
  \centering
  \newcommand\gap{0.495}
  \begin{subfigure}[t]{\gap\linewidth}
      \centering
      \includegraphics[width=\textwidth]{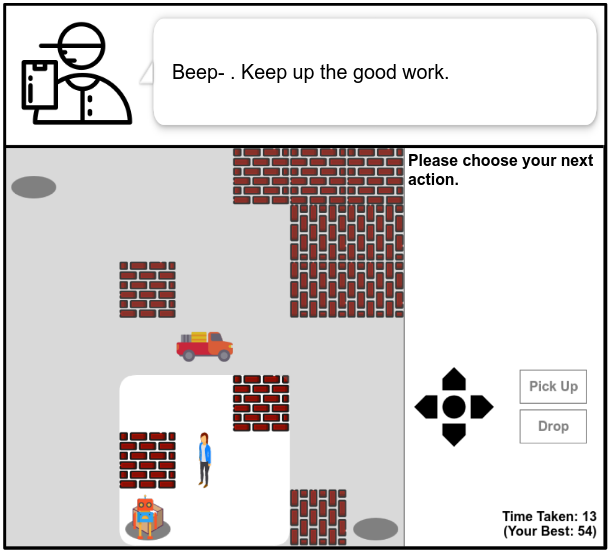}
      \caption{Example: no intervention}
      \label{fig. aicoach no intervention}
  \end{subfigure}
  % \hspace{8ex}
    \hfill
    \begin{subfigure}[t]{\gap\linewidth}
      \centering
      \includegraphics[width=\textwidth]{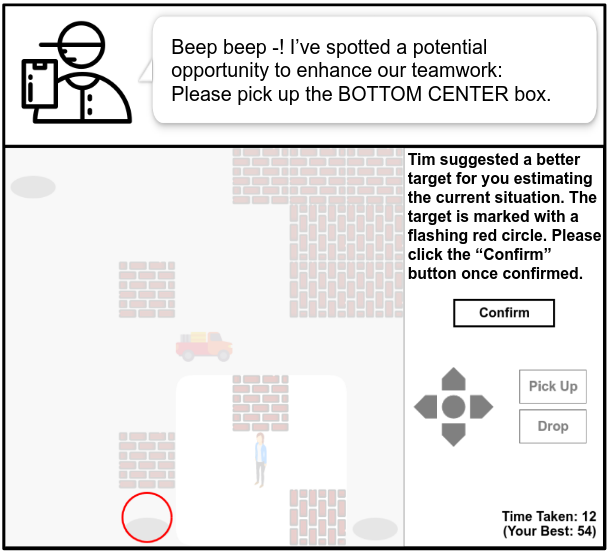}
      \caption{Example: \coach-generated intervention}
      \label{fig. aicoach intervention}
  \end{subfigure} 
  \caption{Snapshots of \coach's interactive user interface (larger images are available in the appendix).}
  \label{fig. aicoach ui}
\end{figure}
\subsection{Execution Phase}
\label{sec. tic execution}

\subsubsection{Intent Detection}
\coach monitors the team during task execution, identifying potential misalignments in team members' intents and computing timely interventions. This capability is enabled by \tic, a framework that has been experimentally shown to generate task-time interventions that enhance teamwork among AI agents~\cite{seo2023automated}. We extend this framework to develop an AI-enabled coaching system for teams that include human members.
During task execution, \coach can observe team members' states and actions, but their intents (a latent variable) remain unobservable. While \coach leverages a human annotator to obtain partial intent annotations during the training phase, involving a human in the loop during task execution is impractical. Therefore, to infer team members' intents, \coach frames the problem as one of Bayesian filtering. Specifically, given the learned model of team behavior $(\mathcal{H}_j \forall j = 1:n)$ and the partial $(s,a)$-trajectory of the team's task execution, \coach employs the forward-backward algorithm to infer each team member's current intent $\hat{x}$. 

\subsubsection{Intervention Generation}
\coach next uses the inferred intents to assess whether the team is aligned. If the intended plans of the team members are likely to lead to suboptimal task performance, \coach intervenes by weighing the costs and benefits of the intervention. Under the \tic framework, determining this balance requires an intervention strategy, which can be hand-crafted or learned. For \coach, we opt for a learned, value-based strategy to minimize human effort in intervention generation. Specifically,\footnote{Since this computation relies on observations, task model, and the learned model of team behavior, it requires no additional human input or domain-specific knowledge.}
 \begin{itemize}
    \item \coach first computes the expected return $(g)$ conditioned on the inferred intent: $g(\hat{x}|s) = E_{\mathcal{H}}[\sum_t \gamma^t r_t |s, \hat{x}]$.
    \item Next, \coach computes the intent values and return for a hypothetical fully aligned team as $x^* = \arg\max_{x} g(x|s)$ and $g(x^*|s) = E_{\pi, \zeta}[\sum_t \gamma^t r_t | s, x^*]$, respectively. We define the benefit of an intervention as the difference between the optimal and estimated return: $g(x^*|s) - g(\hat{x}|s)$.
    \item Finally, if the benefit of an intervention exceeds its cost $c$ by a pre-defined threshold (i.e., $g(x^*|s) - g(\hat{x}|s) > c + \delta$), then \coach prompts the team to pause, reflect on their plans, and recommends the optimal plan corresponding to $x^*$.
\end{itemize}
Choosing an appropriate cost $(c)$ and threshold $(\delta)$ for interventions is crucial, as unnecessary or incorrect interventions could impair team performance and reduce human trust in, and adoption of, \coach. 
\cref{sec. validation} outlines the approach for selecting these hyperparameters for our implementation and evaluation of \coach.

\subsubsection{Intervention Delivery}
To assist human team members, in addition to generating interventions, Socratic requires effective mechanisms for delivering these instructions. In this work, we utilize an interactive user interface for delivering interventions, as illustrated in \cref{fig. aicoach ui} and detailed in \cref{sec: data collection}. Since human team members can choose whether to accept the AI-generated recommendations, Socratic incorporates a hyperparameter $p_a$, which models the probability of a human accepting its recommendation.

\section{Feasibility Studies}
We conducted human subject evaluations to assess the feasibility of AI-enabled coaching in enhancing collaborative task execution. The IRB-approved experimental protocols were designed to evaluate:
\begin{itemize}
    \item[\textbf{Q1.}] Is \coach capable of learning a useful team model?
    \item[\textbf{Q2.}] Is \coach capable of improving team performance?
    \item[\textbf{Q3.}] Is \coach perceived as useful by human users?
\end{itemize}

The evaluations consisted of two studies: training and validation. Both studies involved dyadic teams completing two collaborative tasks. In the training study, we curated a novel dataset of human demonstrations, annotated with intents, to train \coach. In the validation study, we conducted a randomized controlled trial\footnote{Validation Study: The experimental group received coaching from \coach, while the control group completed tasks without any AI-enabled coaching.} to evaluate the objective performance of Socratic and gather subjective feedback from participants regarding AI-enabled coaching. 

\subsection{Domains}
\label{sec. domains}
We first describe the collaborative tasks used in our evaluations: \movers and \rescue. Introduced in \cite{seo2023automated}, these dyadic tasks require teams to maintain a shared plan for effective execution. However, due to partial observability and lack of communication, achieving coordination and high task performance is challenging.

\subsubsection{\movers}
As shown in \cref{fig: movers}, Alice and Rob are tasked with moving three boxes to the truck as quickly as possible. The boxes are heavy and require both teammates to lift them together. Teamwork is effective as long as the teammates agree on which box to move and act accordingly, regardless of the order. However, as depicted in \cref{fig: movers human sight}, each team member has a limited view of the environment and cannot communicate with the other during task execution, making coordination challenging. The task ends after $150$ time steps or when all boxes are moved to the truck, whichever comes first. The cumulative team reward is defined as $150$ minus the time step at which the task terminates. 

\subsubsection{\rescue}
The second task is inspired by time-critical disaster response scenarios. As shown in \cref{fig: rescue}, the environment includes victims at three sites: one at City Hall, two at the Campsite, and four at the Mall. A rescue team, consisting of a police car and a fire truck, must save all victims within a time limit of $30$ time steps. While victims at City Hall and the Campsite can be rescued by a single vehicle, rescuing those at the Mall requires both vehicles to collaborate in repairing one of two bridges. Teamwork in this task is more complex: sometimes the team must work together (e.g., at the Mall), while in other cases, dividing sub-tasks is more efficient (e.g., at City Hall and the Campsite). As depicted in \cref{fig: rescue human sight}, team members can only observe each other when at the same location or a landmark, complicating coordination. The total team reward is defined as the number of victims rescued within the time limit. 

\subsection{Study 1: Training}
\label{sec: data collection}
The first study focused on the training phase of \coach to collect training data and evaluate \textbf{Q1}. Forty participants (20 females, 20 males, mean age: 28.5$\pm$4.9 years) completed the \movers and \rescue tasks with a robot teammate, while also providing annotations of their task-relevant intent $(x \in X)$. For \movers, intent is defined as the box a team member plans to pick up or drop next. For \rescue, intent refers to the site a team member plans to approach next.

\subsubsection{Materials and Setup}
We developed a website using the Flask framework~\cite{grinberg2018flask} that included the two tasks, complete with a user interface for task execution and intent labeling (\cref{fig. ui}). This platform enabled participants to perform the experiment remotely. Each participant was paired with a robot teammate, forming a dyadic human-robot team. Following \cref{sec. teamwork model}, behavior of each teammate was modeled as  $\mathcal{H}_j = (X, \pi_j, \zeta_j; \mathcal{M})$. The robot (denoted as $R$) had its policy $\pi_R$ pre-trained using value iteration, and its intent dynamics $\zeta_R$ were manually specified. The experiment aimed to collect data on the human teammate's (denoted as $H$) behavior in order to learn their policy $\pi_H$ and intent dynamics $\zeta_H$. Both teammates had to make decisions under partial observability and infer the intent of their teammate to complete the task successfully.

\subsubsection{Procedure}
\label{sec: data collection procedures}
Upon providing informed consent, participants were introduced to the experiment and completed a demographic survey. They were then instructed to complete the dyadic tasks with the robot, following the same process for both \movers and \rescue. This process included an interactive tutorial and four task trials. The tutorial introduced participants to the task and trained them on how to navigate the user interface (UI). The tutorial featured a guided scenario that mirrored the actual task.
For each domain, participants proceeded through four task trials after completing the tutorial. Each trial was followed by a simplified \textit{after-action review}~\cite{morrison1999foundations, taberski2021visualizing, qian2024measuring}. 
During each trial, the website displayed a task scene and a task control UI, allowing participants to control their character to complete the task (\cref{fig. ui for task}). The experiment collected data on task states $(s)$ and team actions $(a)$ while generating human intent annotations $(x)$. Intent annotations were generated during the task and refined via the after-action reviews, as described in \cref{sec: annotation}. After completing four trials for both the \movers and \rescue tasks, the experiment concluded with a post-experiment survey, where participants provided open-ended feedback about their experience.

\subsubsection{Annotation}
\label{sec: annotation}
Training \coach requires both observable $(s,a)$-trajectories and time series data of team members' intents $(x)$, which are latent and must be manually annotated. In this study, we collected intent data through participant reports, supported by user-centered annotation mechanisms to ensure reliable data collection.
Recall that in both domains, intent is tied to a physical location in the task scene, such as a box or a rescue site. To streamline reporting, we developed a ``Destination Selection'' UI, allowing participants to report their intended destination during task execution (\cref{fig. ui for intent selection}). Potential destinations are highlighted, and participants select their intended location with a mouse click. Participants are encouraged to update their intent when it changes and are prompted if five time steps pass without a report. Selected intents are visually indicated with a flashing red circle. Additionally, key actions like ``Pick Up,'' ``Drop,'' or ``Rescue'' are restricted to the selected destination, ensuring alignment between reported intents and actions. After each task trial, participants use the "after-action review" UI to verify and, if needed, correct their annotations (\cref{fig. ui for post-session review}). This interface replays the task execution, displaying both team actions and selected intents, allowing participants to confirm their reports. If discrepancies are found, participants can adjust incorrect intents using the ``Fix Destination'' button, improving the accuracy of the dataset used to train and validate \coach.

\subsubsection{Data Analysis}
\label{sec. learning team behavior model}
We collected 160 demonstrations per domain and trained \coach using a semi-supervised approach. Recognizing that intent annotation is resource-intensive, we used only 30\% of the intent labels for training and reserving the rest for validation. This approach enables evaluating \coach in a more realistic setting, where only partial intent annotations are available.
\begin{table}
\caption{Survey Statements}
\label{table: survey}
\begin{center}
\small
\setlength\dashlinegap{3pt}
\begin{tabular}{cl} \toprule
\# & Statement (rated on a 5-point Scale) \\ \midrule
1 & \makecell[tl]{The team worked fluently together.} \\
2 & \makecell[tl]{The robot contributed to the fluency of the interaction.} \\
3 & \makecell[tl]{The team improved over time.}  \\ \midrule %\hdashline \noalign{\vskip 0.5ex}
4 & \makecell[tl]{During the task, I followed the AI Coach suggestions in general.} \\
5 & \makecell[tl]{The AI Coach was intelligent.} \\
6 & \makecell[tl]{The AI Coach was trustworthy.} \\
7 & \makecell[tl]{The AI Coach's suggestions were effective.} \\
8 & \makecell[tl]{The AI Coach's suggestions were timely.} \\
9 & \makecell[tl]{The AI Coach contributed to the fluency of the interaction.} \\
\bottomrule  
\end{tabular} 
\end{center}
\end{table}
\subsection{Study 2: Validation}
\label{sec. validation}
After collecting the training data, we conducted a second study to evaluate \coach's performance (\textbf{Q2}) and perceived usefulness (\textbf{Q3}). The study was a randomized control trial, where only the experimental group received coaching from \coach. %The control group completed tasks without AI-enabled coaching.

\subsubsection{Participants}
We recruited participants via Prolific~\cite{palan2018prolific}. Of the 73 users who accessed the experiment, 61 completed it. To ensure balanced group sizes, we used the first 30 participants from each group. The control group consisted of 13 females and 17 males (age: 28.7$\pm$8.4 years), while the experimental group included 11 females, 17 males, and 2 non-binary participants (27.8$\pm$9.1 years).

\subsubsection{Materials and Setup}
Similar to the first study, we developed a website featuring the two tasks with an interactive user interface. However, instead of intent annotation mechanisms, this version incorporated \coach on the backend and its user interface on the frontend for interacting with the team during task execution. As shown in \cref{fig. aicoach ui}, the interface features an AI coach icon with a speech balloon above the task screen.
During task execution, the speech balloon nominally displays: \textit{``Keep up the good work.''}
However, if \coach detects misaligned intents and decides to intervene, it pauses the task, prompts the team to reflect on their plans, and recommends an optimal course of action corresponding to $x^*$. As illustrated in \cref{fig. aicoach intervention}, the speech balloon displays:
\begin{quote}
\centering
\textit{``I've spotted a potential opportunity to enhance our teamwork: Please $\langle recommendation\rangle$''}. 
\end{quote}
The suggestion is highlighted with a red circle, and participants must click the ``Confirm'' button to resume the task.
While \coach offers recommendations, participants ultimately decide whether to accept them.
To account for the fact that not all recommendations will be followed, we set $p_a = 0.9$ for this study. 
\coach utilizes two additional hyperparameters: the cost of intervention $c$ and the threshold $\delta$. For \movers, the intervention cost is set to 1, representing the loss of one time step to pause and reflect on the recommendation. In contrast, for the life-critical \rescue task, the cost is considered negligible ($c = 0$) as any small delays caused by interventions are justified if they assist in rescue efforts. The threshold $\delta$ was determined through a grid search over the hyperparameter space, with values set to $5$ for \movers and $0.1$ for \rescue based on simulated experiments with the learned teamwork model.

\begin{table}
\centering
\caption{Success Rate of the Learned Model}%
\label{table: btil performance}
\setlength\dashlinegap{4pt}
\small
\begin{tabular}{@{}c@{}c@{\hskip 6pt}c@{\hskip 6pt}c@{\hskip 6pt}c@{}} \toprule
~~Domain~~ & ~~Intent~~ & \makecell[c]{Success (\%)} & \makecell[c]{Wrong (\%)} & \makecell[c]{Nowhere (\%)} \\ \midrule
\multirow{5}{*}{\small \movers} & Box 1 & 75.4 & 5.9 & 18.7  \\
                                & Box 2 & 72.5 & 12.5 & 15.0  \\
                                & Box 3 & 69.3 & 13.8 & 16.9  \\
                                & Truck & 99.6 & 0.0 & 0.4  \\ \cmidrule{2-5}
                                & Mean & 79.2 & 8.05 & 12.75  \\ \midrule
\multirow{5}{*}{\small \rescue} & City Hall & 88.3 & 0.3 & 11.4  \\
                                & Campsite & 38.2 & 9.2 & 52.6  \\
                                & Bridge 1 & 59.4 & 2.2 & 38.4  \\
                                & Bridge 2 & 45.4 & 3.1 & 51.5  \\ \cmidrule{2-5}
                                & Mean & 57.8 & 3.7 & 38.5  \\ \bottomrule
\end{tabular} 
\end{table}
\subsubsection{Procedure}
The overall structure of this experiment closely mirrors that of the first study. It is web-based and includes a study overview, a demographic survey, \movers and \rescue domains, and a post-experiment survey. For each domain, participants completed an interactive tutorial followed by four task trials. While the tutorials and trials were similar to the first study, intent annotation features were removed. Only for the experimental group, \coach's features were integrated into the tutorial and task trials. Each domain involved one practice trial to help participants familiarize themselves with the task and the robot teammate, followed by three test trials. Neither group received assistance from \coach during the practice trial. In the test trials, the control group performed the task without coaching, while the experimental group received task-time interventions from \coach. After the trials, participants completed the survey described next.

\subsubsection{Measures}
We assess \textbf{Q1} by quantifying the intent-condition-ed success rate of the learned model. For \textbf{Q2}, team performance is evaluated using task scores. Beyond improving teamwork, the perceived usefulness of \coach is essential for its adoption by human users. Hence, to address \textbf{Q3}, we use subjective statements adapted from a widely used scale~\cite{hoffman2019evaluating}. The first three questions solicited participants perception regarding the robot teammate, while the rest regarding \coach the AI coach. Control group rated the first three statements listed in \cref{table: survey}, while the experimental group rated all statements. Responses were recorded on a 5-point scale, ranging from \textit{strongly disagree} (1) to \textit{strongly agree} (5).

% \section{Findings}
% \label{sec:findings}
% We conclude by discussing the experimental results and highlighting their key implications for both team training and AI research.
\subsection{Experimental Results}
\label{sec:findings}

\subsubsection{\coach learns intent-driven models of team behavior.}
\label{sec: model learning results}
To address \textbf{Q1}, \coach first learns models of team behavior using the training data. We then evaluate if the learned model captures intent-driven behaviors by simulating the policy $1000$ times for each intent $x$ and measuring its success rate in completing the intended sub-task within 20 time steps. For instance, if the specified intent is to rescue victims at City Hall, we check how often the model succeeds.
\cref{table: btil performance} presents the success rates of the learned model for each intent. Failures are categorized as either \textit{Wrong} (where the model accomplishes a sub-task associated with a different intent, such as rescuing victims at the Camp Site when the specified intent was City Hall) or \textit{Nowhere} (where the model fails to complete any sub-task within the time limit). 
On the challenging task of modeling team behaviors from human data, the model achieved an average success rate of 79\% for \movers and 58\% for \rescue. Most failures belong to the \textit{Nowhere} category, suggesting that model learns intent-driven models of team behavior. Through the second study, we find that this model learning performance is sufficient for \coach to deliver effective task-time interventions to improve teamwork.

\begin{figure}
\centering
\includegraphics[width=\linewidth]{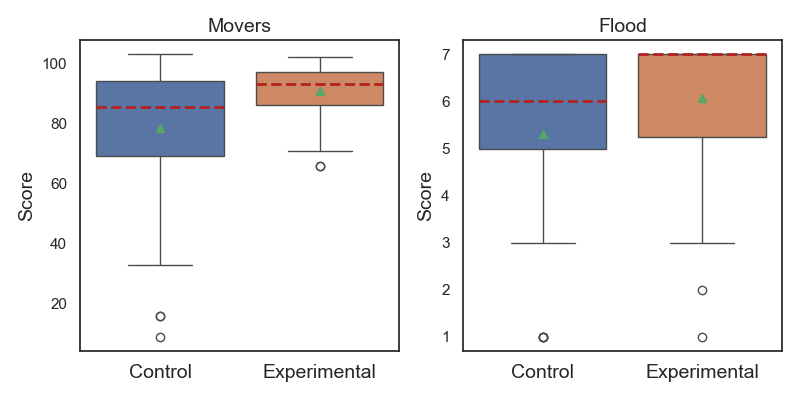} \vspace{-1em}
\caption{Team Scores: with and without \coach.} \vspace{-1em}
\label{fig. task results} 
\end{figure}
\subsubsection{\coach improves teamwork via targeted interventions.}
To answer \textbf{Q2}, we compared the performance of the two groups, using a cost-adjusted score that accounts for the time spent on processing and responding to interventions. Specifically, \score is defined as $R - C$, where $R$ is the cumulative team reward and $C$ is the total cost of interventions. For the control group, \score is equal to the task score, as no interventions took place.
As shown in \cref{fig. task results}, the experimental group outperformed the control group. In \movers, teams coached by \coach scored on average $90.8 (\pm 8.4)$ compared to $78.6 (\pm 21.7)$ for the control group, with a statistical significance of $p < 0.001$. Similarly, for \rescue, the average score of the experimental group was $6.1 (\pm 1.5)$ versus $5.3 (\pm 1.7)$ for the control group, with $p < 0.01$. The average number of interventions was 3.9 ($\pm$1.7) for \movers and 2.3 ($\pm$2.1) for \rescue. These results highlight that \coach effectively enhanced teamwork with minimal interventions. This targeted approach to delivering interventions is crucial for improving team performance and building human users' trust in AI-enabled coaching.

\subsubsection{\coach is perceived as useful by human users.}
\label{sec: survey results}
\begin{figure}
  \centering
  \includegraphics[width=\linewidth]{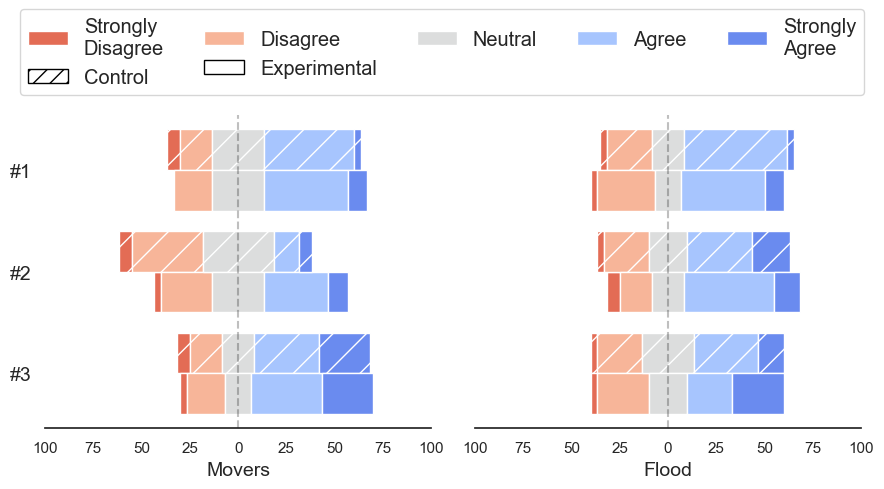}
  \newline
  \includegraphics[width=\linewidth]{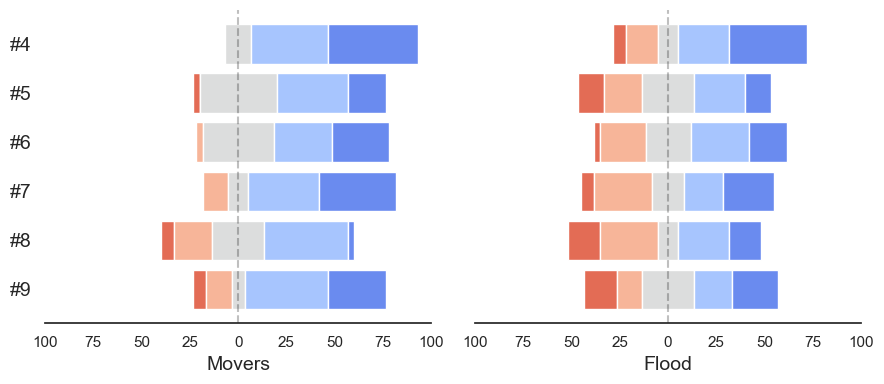} \vspace{-1em}
  \caption{Participant Responses to Survey Statements}
  \label{fig: survey plot} 
\end{figure}

To answer \textbf{Q3}, we analyzed participants' survey responses. \cref{fig: survey plot} displays the percentage of positive, neutral, and negative assessments for each statement. 
Responses to statements \#1-3, which evaluated the robot teammate, were largely similar across both groups, indicating that participants had comparable perceptions of the robot teammate's capabilities. This consistency ensures a fair comparison between the groups, allowing us to accurately evaluate the AI coach's utility.

Statements \#4-9, which evaluated \coach and were rated only by the experimental group, indicate that participants perceived \coach as useful, effective, intelligent, and trustworthy.
Based on these statements, the average rating of \coach was $3.81 (\pm 1.03)$ for \movers and $3.28 (\pm 1.32)$ for \rescue on a $1-5$ scale.
Except for statement \#8 for \rescue task, the positive responses outweighed the negative ones for all statements.

Open-ended feedback suggested that \coach was seen as more helpful in the \movers task, while participants found \rescue more challenging. 
Regarding statement \#8, which asked about the timeliness of \coach’s recommendations, one participant commented:
\begin{quote}
\textit{“There was one occasion when the AI’s suggestion came a bit late, causing me to waste a few moves.”}
\end{quote}
While \coach is already designed to provide proactive guidance using a predictive model of teamwork, participants' responses suggest that they value this proactivity and may expect even more planning support from an AI Coach.  Informed by these findings, we conclude by summarizing our contributions and discussing their implications for both team training and AI research.

\section{Conclusion}
\label{sec: conclusion}
We introduce \coach: a system that provides AI-enabled coaching to teams with human members during task execution. Through human subject experiments on challenging dyadic tasks, we demonstrated that \coach not only enhances team performance but is also perceived as useful by participants. Since \coach does not perform the tasks itself, it has the potential to assist in various domains, including those where AI agents may lack the capability to act but can still analyze and enhance human task execution.

Along with its strengths, we also highlight the limitations of this proof-of-concept work, which suggest exciting future research directions. First, while the experimental tasks captured challenging elements of real-world collaboration, they were conducted in a web-based environment. Future work should investigate AI-enabled coaching in more complex scenarios that include dynamic environments, larger teams, multiple objectives, ad-hoc collaboration, or members with diverse expertise~\cite{stone2010ad, seo2024idil, seo2025hierarchical}. Additionally, expanding AI coaching to physical teamwork settings using multimodal perception is an important next step~\cite{wang2019ai, xia2025sportu}.

Second, from a human-centered perspective, our study opens up new opportunities to examine how AI can support team training and complement human coaches. Informed by the science of teamwork, expanding \coach's interventions to include more varied recommendations would further enhance its utility~\cite{peters2013team, weaver2014team, britton2015expanding}. To enhance usability, interventions could be delivered through user-friendly interfaces such as screens, audio systems, or augmented/virtual reality.
Lastly, given AI coaches can make errors, a critical area for further investigation is ensuring the safe and responsible deployment of AI coaches~\cite{hoffman2013trust,  yang2017evaluating, qian2022evaluating, quintero2023robotic, qian2024pps, meneses2024perceptions}. This includes examining how trust in AI coaches can be effectively built, calibrated, and maintained to foster successful human-AI collaboration.

%%%%%%%%%%%%%%%%%%%%%%%%%%%%%%%%%%%%%%%%%%%%%%%%%%%%%%%%%%%%%%%%%%%%%%%%

%%% The acknowledgments section is defined using the "acks" environment
%%% (rather than an unnumbered section). The use of this environment 
%%% ensures the proper identification of the section in the article 
%%% metadata as well as the consistent spelling of the heading.

\begin{acks}
% We thank the anonymous reviewers.
This research was supported by NSF award $\#2205454$.
\end{acks}

\ifarxiv
\appendix

\section{Snapshots of Human Experiments.}
Figs. \ref{fig. larger UI for task}-\ref{fig. larger aicoach intervention} are larger versions of Figs. \ref{fig. ui}-\ref{fig. aicoach ui} mentioned in the main text.

\newcommand\gap{0.92}
\begin{figure}[h]
  \centering
  \includegraphics[width=\gap\linewidth]{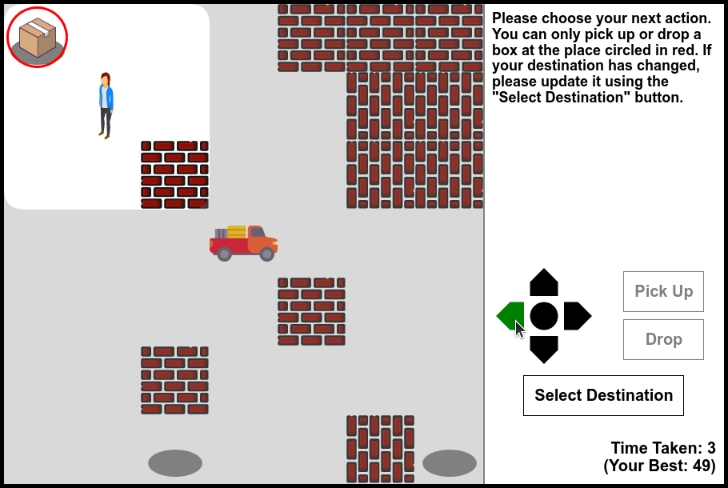}
  \caption{Task scene and UI to control Alice (Fig. \ref{fig. ui for task}).}
  \label{fig. larger UI for task}
\end{figure}
  
\begin{figure}[h]
  \centering
  \includegraphics[width=\gap\linewidth]{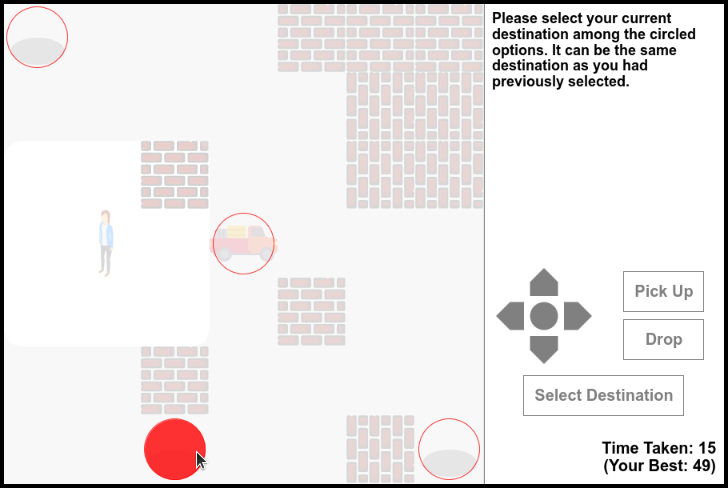}
  \caption{UI for intent annotation (Fig. \ref{fig. ui for intent selection}).}
  \label{fig. larger UI for intent selection}
\end{figure}

\begin{figure}[h]
  \centering
  \includegraphics[width=\gap\linewidth, frame]{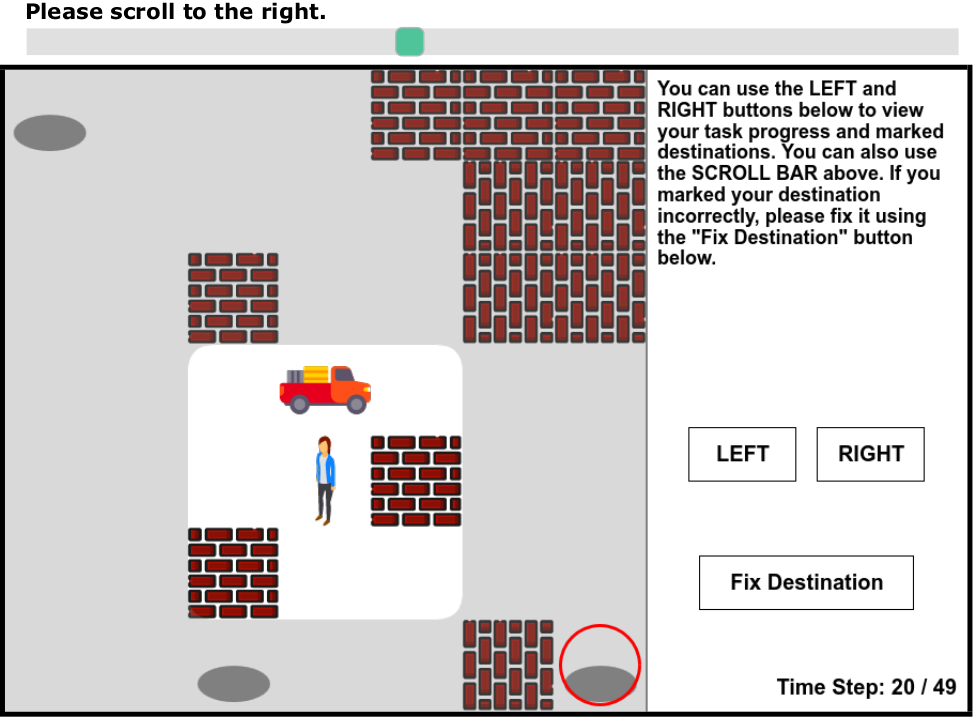}
  \caption{UI for after-action review (Fig. \ref{fig. ui for post-session review}).}
  \label{fig. larger UI for post-session review}
\end{figure}

\begin{figure}[h]
  \centering
  \includegraphics[width=\gap\linewidth]{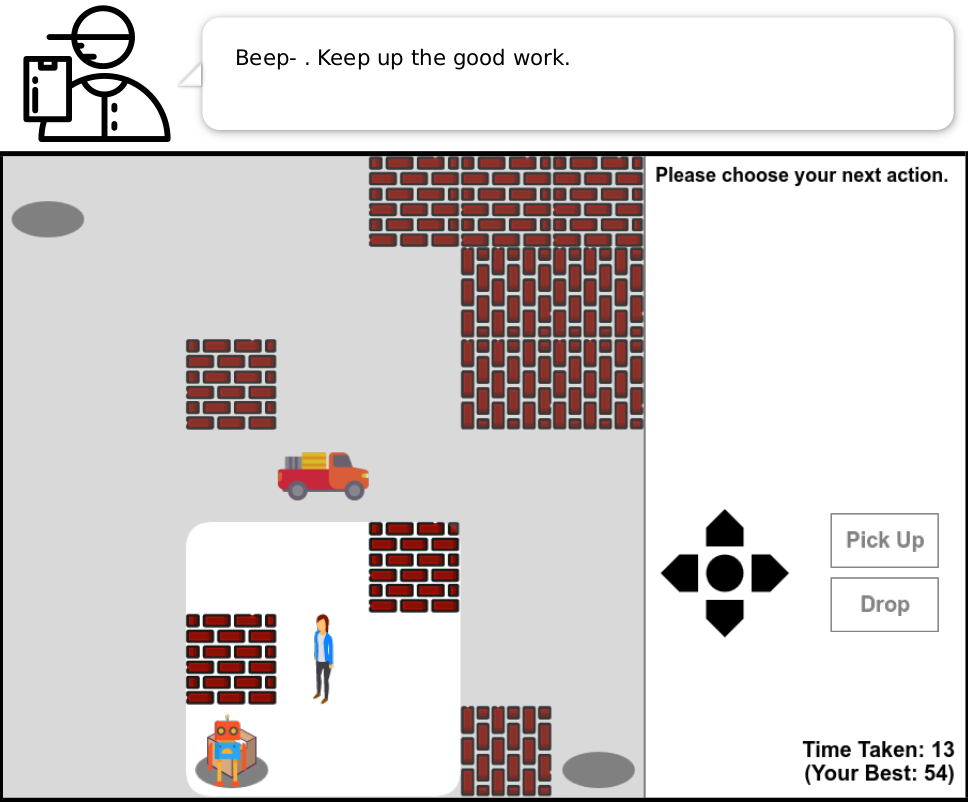}
  \caption{AI coach: no intervention (Fig. \ref{fig. aicoach no intervention}).}
  \label{fig. larger aicoach no intervention}
\end{figure}

\begin{figure}[h]
  \centering
  \includegraphics[width=\gap\linewidth]{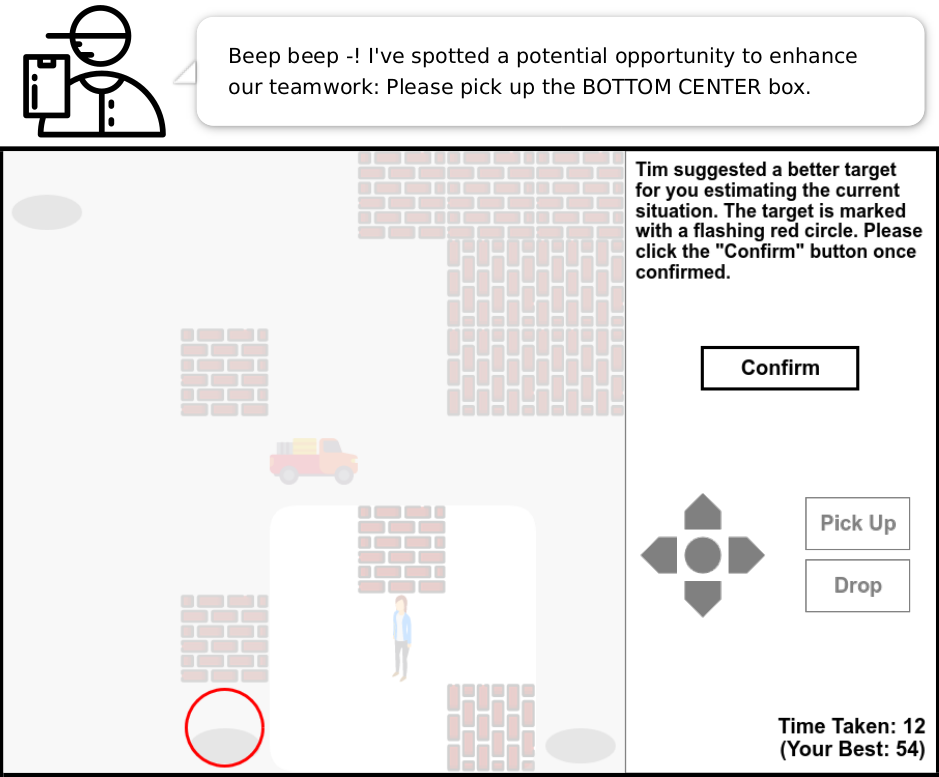}
      \caption{AI coach: \coach-generated intervention (Fig. \ref{fig. aicoach intervention}).}
  \label{fig. larger aicoach intervention}
\end{figure}

\section{Additional Discussion on the results in Section \ref{sec: model learning results}}
As demonstrated in Table \ref{table: btil performance}, the success rates differ between the intents. This difference is due to the varying characteristics of the environments and the degree of heterogeneity across demonstrations. For instance, when a human is carrying a box, they are almost always heading toward the truck. This enables the learned model to clearly associate the intent "truck" with the behavior of moving toward the truck. In contrast, when a human is not carrying a box, they could be heading toward any of Box 1, Box 2, or Box 3. This ambiguity makes intent inference for Box 1, Box 2, and Box 3 more challenging than intent inference for "truck."

\section{Statistical Analysis of survey results in Section \ref{sec: survey results}}

Statements \#1-3 pertained to the robot teammate and were presented to both groups. Statements \#4-9 focus on the AI Coach and were asked only of the experimental group.
Using a scale of 1-5 to represent "Strongly Disagree" to "Strongly Agree," we performed statistical analysis for statements \#1-3. As shown in Table \ref{table: survey stats}, participant's responses were not statistically significant between the control and experimental groups. This suggests that participants from both groups had similar impressions of their robot teammate, indicating that no confounds were introduced by the robot teammate.

\begin{table}[h]
    \caption{Statistical Analysis for Survey Statements \#1-3.}
    \label{table: survey stats}
    \centering
    \begin{tabular}{c c c c c}
        \toprule
        \textbf{Domain} & \# & \textbf{Control} & \textbf{Experimental} & \textbf{p-value} \\
        \midrule
        \multirow{3}{*}{Movers} 
         & 1 & 3.23 $\pm$ 1.01 & 3.43 $\pm$ 0.94 & 0.43 \\
         & 2 & 2.77 $\pm$ 1.01 & 3.20 $\pm$ 1.06 & 0.11 \\
         & 3 & 3.57 $\pm$ 1.25 & 3.63 $\pm$ 1.19 & 0.83 \\
        \midrule
        \multirow{3}{*}{Flood} 
         & 1 & 3.30 $\pm$ 0.99 & 3.27 $\pm$ 1.11 & 0.90 \\
         & 2 & 3.43 $\pm$ 1.17 & 3.43 $\pm$ 1.14 & 1.00 \\
         & 3 & 3.30 $\pm$ 1.09 & 3.43 $\pm$ 1.25 & 0.66 \\
        \bottomrule
    \end{tabular}
\end{table}

For responses regarding the AI Coach usefulness (statements \#4-9), using combined scores rather than individual scores is recommended \cite{schrum2020four}. Overall, the \% of positive responses was 77.8\% for Movers and 57.8\% for Flood. Fig. \ref{fig: survey plot} illustrates this distribution, with the right side of the vertical line being more prominent.

\fi

% \newpage

%%%%%%%%%%%%%%%%%%%%%%%%%%%%%%%%%%%%%%%%%%%%%%%%%%%%%%%%%%%%%%%%%%%%%%%%

%%% The next two lines define, first, the bibliography style to be 
%%% applied, and, second, the bibliography file to be used.


%%%%%%%%%%%%%%%%%%%%%%%%%%%%%%%%%%%%%%%%%%%%%%%%%%%%%%%%%%%%%%%%%%%%%%%%

\end{document}